\documentclass[journal]{IEEEtran}
\usepackage{slashbox,graphicx,times,amsmath,amssymb,cite,subfigure,stfloats,booktabs, url,multirow,flushend}
\usepackage[lined,ruled,commentsnumbered]{algorithm2e}

\makeatletter

\newcommand{\Rmnum}[1]{\expandafter\@slowromancap\romannumeral #1@}
\makeatother

\begin{document}
\title{On the Vulnerability of CNN Classifiers\\ in EEG-Based BCIs}

\author{
    Xiao~Zhang and Dongrui~Wu
    \thanks{
        X.~Zhang and D.~Wu are with the Key Laboratory of Image Processing and Intelligent Control (Huazhong University of Science and Technology), Ministry of Education. They are also with the School of Artificial Intelligence and Automation, Huazhong University of Science and Technology, Wuhan 430074, China. Email: xiao\_zhang@hust.edu.cn, drwu@hust.edu.cn.
    }
    \thanks{
        Dongrui~Wu is the corresponding author.
    }
}

\maketitle

\begin{abstract}
Deep learning has been successfully used in numerous applications because of its outstanding performance and the ability to avoid manual feature engineering. One such application is electroencephalogram (EEG) based brain-computer interface (BCI), where multiple convolutional neural network (CNN) models have been proposed for EEG classification. However, it has been found that deep learning models can be easily fooled with adversarial examples, which are normal examples with small deliberate perturbations. This paper proposes an unsupervised fast gradient sign method (UFGSM) to attack three popular CNN classifiers in BCIs, and demonstrates its effectiveness. We also verify the transferability of adversarial examples in BCIs, which means we can perform attacks even without knowing the architecture and parameters of the target models, or the datasets they were trained on. To our knowledge, this is the first study on the vulnerability of CNN classifiers in EEG-based BCIs, and hopefully will trigger more attention on the security of BCI systems.
\end{abstract}


\IEEEpeerreviewmaketitle

\section{Introduction} \label{sect:Intro}

    A brain-computer interface (BCI) is a communication pathway between the human brain and a computer~\cite{BCIIntro}. Electroencephalogram (EEG), which measures the brain signal from the scalp, is the most widely used input signal in BCIs, due to its simplicity and low cost~\cite{BCIReview}. There are different paradigms in using EEG signals in BCIs, e.g., P300 evoked potentials~\cite{P300Sutton, P3001988,drwuTHMS2017,drwuTNSRE2016}, motor imagery (MI)~\cite{MI2001}, steady-state visual evoked potential (SSVEP)~\cite{SSVEPSurvey}, drowsiness/reaction time estimation \cite{drwuSF2018,drwuTFS2017,drwuRG2017}, etc.

\begin{figure}[htbp]         \centering
        \includegraphics[width=\linewidth,clip]{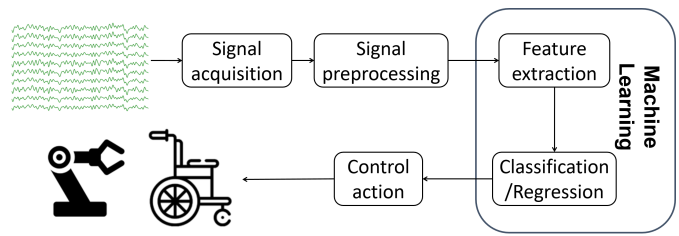}\\
    \caption{The procedure of a BCI system when traditional machine learning algorithms are used. Manual feature extraction is not necessary if deep learning is employed.} \label{fig:procedure}
    \end{figure}

    As shown in Fig.~\ref{fig:procedure}, a BCI system usually consists of four parts: \emph{signal acquisition}, \emph{signal preprocessing}, \emph{machine learning}, and \emph{control action}. The machine learning block usually includes feature extraction and classification/regression if traditional machine learning algorithms are applied. However, feature extraction and classification/regression can also be seamlessly integrated into a single deep learning model.

    Deep learning has achieved state-of-the-art performance in various applications, without the need of manual feature extraction. Recently, multiple deep learning models, particularly those based on convolutional neural networks (CNNs), have also been proposed for EEG classification in BCIs. Lawhern et al. (2016)~\cite{EEGNet} proposed EEGNet, which demonstrated outstanding performance in several BCI applications. Schirrmeister et al. (2017)~\cite{MNE} designed a deep CNN model (DeepCNN) and a shallow CNN model (ShallowCNN) to perform end-to-end EEG decoding. There were also some works converting EEG signals to images and then classifying them with deep learning models~\cite{EEG2Image, DeepLearningMI, Tayeb2019}. This paper focuses on the CNN models that accept the raw EEG signal as the input, more specifically, EEGNet, DeepCNN, and ShallowCNN. A CNN model using the spectrogram as the input is briefly discussed in Section~\ref{sect:Discussion}.

    Albeit their outstanding performance, deep learning models are vulnerable to adversarial attacks. In such attacks, deliberately designed small perturbations, many of which may be hard to notice even by human, are added to normal examples to fool the deep learning model and cause dramatic performance degradation. This phenomena was first investigated by Szegedy et al. in 2013~\cite{AdvExamSzegedy} and soon received great attention. {Goodfellow et al. (2014)}~\cite{FGSM} successfully confused a deep learning model to misclassify a panda into a gibbon. {Kurakin et al. (2016)}~\cite{BIM} found that deep learning systems might even make mistakes on printed photos of adversarial examples. {Brown et al. (2017)}~\cite{AdvPatch} made an adversarial patch which was able to confuse deep learning models when attached on a picture. {Athalye et al. (2017)}~\cite{Adv3D} built an adversarial 3D-printed turtle which was classified as a riffle at every viewpoint with randomly sampled poses. Additionally, adversarial examples have also been used to attack speech recognition systems, e.g., a piece of speech, which is almost the same as a normal one but with a small adversarial perturbation, could be transcribed into any phrase the attacker chooses~\cite{AdvExamAudio}.

    Adversarial examples could significantly damage deep learning models, which have been an indispensable component in computer vision, automatic driving, natural language processing, speech recognition, etc. To our knowledge, the vulnerability of deep learning models in EEG-based BCIs has not been investigated yet, but it is critical and urgent. For example, EEG-based BCIs could be used to control wheelchairs or exoskeleton for the disabled \cite{Li2016}, where adversarial attacks could make the wheelchair or exoskeleton malfunction. The consequence could range from merely user confusion and frustration, to significantly reducing the user's quality of life, and even to hurting the user by driving him/her into danger on purpose. In clinical applications of BCIs in awareness evaluation/detection for disorder of consciousness patients \cite{Li2016}, adversarial attacks could lead to misdiagnosis.

    According to how much the attacker can get access to the target model, attacks can be categorized into three types:
    \begin{enumerate}
    \item \emph{White-box attacks}, which assume that the attacker has access to all information of the target model, including its architecture and parameters. Most of the white-box attacks are based on some optimization strategies or gradient strategies, such as {L-BFGS}~\cite{AdvExamSzegedy}, DeepFool~\cite{Moosavi-Dezfooli2016}, the C\&W method~\cite{AdvCW}, the fast gradient sign method (FGSM)~\cite{FGSM}, the {basic iterative method}~\cite{BIM}, etc.
    \item \emph{Black-box attacks}, which assume the attacker knows neither the architecture nor the parameters of the target model, but can observe its responses to inputs. {Papernot et al. (2016)}~\cite{BlackBoxAttack} developed a black-box attack approach which can be used to generate adversarial examples by interacting with the target model and training a substitute model. {Su et al. (2017)}~\cite{OnePixelAttack} successfully fooled three different models by changing just one pixel of an image. {Brendel et al. (2017)}~\cite{DecisionBasedAttack} proposed a black-box attack approach that starts from a large adversarial perturbation and then tries to reduce the perturbation while staying adversarial.
    \item \emph{Gray-box attacks}, which assume the attacker knows some but not all information about the target model, e.g., the training data that the target model is tuned on, but not its architecture and parameters.
    \end{enumerate}

   In order to better compare the application scenarios of the three attack types, we summarize their characteristics in Table~\ref{tab:summary}. `$-$' means that whether the information is available or not will not affect the attack strategy, since it will not be used in the attack. It is clear that we need to know less and less information about the target model when going from white-box attack to gray-box attack and then to black-box attack. This makes the attack more and more practical, but we would also expect that knowing less information about the target model may affect the attack performance. This paper considers all three types of attacks in EEG-based BCIs.

        \begin{table}[htbp] \center
        \caption{Summary of the three attack types.}   \label{tab:summary}
        \begin{tabular}{c|ccc}
        \toprule
        Target model information& White-Box & Gray-Box & Black-Box \\ \midrule
        Know its architecture & $\checkmark$ & $\times$ & $\times$ \\
        Know its parameters $\boldsymbol{\theta}$ & $\checkmark$ & $\times$ & $\times$ \\
        Know its training data &  $-$ & $\checkmark$ & $\times$ \\
        Can observe its response &  $-$ & $-$ & $\checkmark$ \\
        \bottomrule
        \end{tabular}
        \end{table}

    According to its purpose, an attack can also be regarded as a \emph{target attack}, which forces a model to classify an adversarial example into a specific class, or a \emph{nontarget attack}, which only forces a model to misclassify the adversarial examples.

\begin{figure}[htbp]         \centering
        \includegraphics[width=\linewidth,clip]{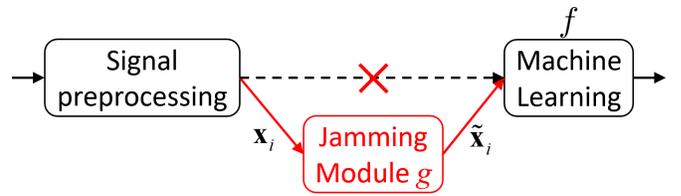}
    \caption{Our proposed attack framework: inject a jamming module between signal preprocessing and machine learning to generate adversarial examples.} \label{fig:injection}
    \end{figure}

\begin{figure}[htbp]         \centering
        \includegraphics[width=\linewidth,clip]{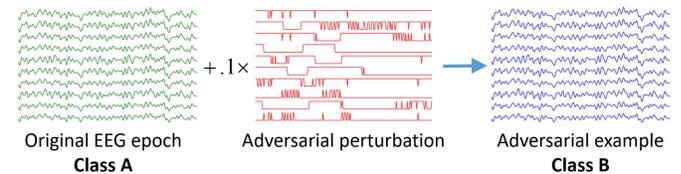}
    \caption{A normal EEG epoch and its adversarial example. The two inputs could be classified into different classes, although they are almost identical. } \label{fig:IntroExample}
    \end{figure}

    This paper aims at exploring the vulnerability of CNN classifiers under nontarget adversarial examples in EEG-based BCIs. We propose an attack framework which converts a normal EEG epoch into an adversarial example by simply injecting a jamming module before machine learning to perform adversarial perturbation, as shown in Fig.~\ref{fig:injection}. We then propose an unsupervised fast gradient sign method (UFGSM), an unsupervised version of FGSM \cite{FGSM}, to design the adversarial perturbation. The adversarial perturbation could be so weak that it is hardly noticeable, as shown in Fig.~\ref{fig:IntroExample}, but can effectively fool a CNN classifier. We consider three attack scenarios -- white-box attack, gray-box attack, and black-box attack -- separately. For each scenario, we provide the attack strategy to craft adversarial examples, and the corresponding experimental results. We show that our approaches can work in most cases and can significantly reduce the classification accuracy of the target model. To our knowledge, this is the first study on the vulnerability of CNN classifiers in EEG-based BCIs. It exposes an important security problem in BCI, and hopefully will lead to the design of safer BCIs.

    The remainder of the paper is organized as follows: Section~\ref{sect:AdvExam} proposes the strategies we use to attack the CNN classifiers. Section~\ref{sect:experiments} presents the details of the experiments and the results on white-box attack, gray-box attack, and black-box attack. Section~\ref{sect:conclusions} draws conclusion and points out some future research directions.

\section{Strategies to Attack BCI Systems} \label{sect:AdvExam}

    This section introduces our strategies to attack EEG-based BCI systems when CNN is used as the classifier. We assume the attacker is able to invade a BCI system and inject a jamming module between signal preprocessing and machine learning, as shown in Fig.~\ref{fig:injection}. This is possible in practice, as many BCI systems transmit preprocessed EEG signals to a computer, a smart phone, or the cloud, for feature extraction and classification/regression. The attacker could attach the jamming module to the EEG headset signal transmitter, or to the receiver at the other side, to perform adversarial perturbation.

    The jamming module needs to satisfy two requirements: 1) the adversarial perturbation it generates should be so small that it is hardly detectable; and, 2) the adversarial example can effectively fool the CNN classifier. We propose UFGSM to construct it.

    Let
    \begin{align}
    \mathbf{x}_i=\left[\begin{array}{ccc}
                         \mathbf{x}_i(1,1), & \cdots, & \mathbf{x}_i(1,T) \\
                         \vdots & \ddots & \vdots \\
                         \mathbf{x}_i(C,1), & \cdots, & \mathbf{x}_i(C,T)
                       \end{array}\right]
    \end{align}
    be the $i$-th raw EEG epoch, where $C$ is the number of EEG channels and $T$ the number of the time domain samples. Let $y_i$ be the true class label associated with $\mathbf{x}_i$, $f(\mathbf{x}_i)$ the mapping from $\mathbf{x}_i$ to $y_i$ used in the target CNN model, $\widetilde{\mathbf{x}}_i=g(\mathbf{x}_i)$ the adversarial perturbation generated by the jamming module $g$. Then, $g$ needs to satisfy:
    \begin{align}
        &|\widetilde{\mathbf{x}}_i(c,t)-\mathbf{x}_i(c,t)| \leqslant \epsilon,\quad \forall c\in[1,C],\  t\in[1,T] \label{eq:limitation}\\
        &f(\widetilde{\mathbf{x}}_i) \neq y_i \label{eq:target}
    \end{align}
    Equation (\ref{eq:limitation}) ensures that the perturbation is no larger than a predefined threshold $\epsilon$, and (\ref{eq:target}) requires the generated adversarial example should be misclassified by the target model $f$. Note that (\ref{eq:limitation}) must hold for every $c\in[1,C]$ and $t\in[1,T]$, but it may be impossible for every $\widetilde{\mathbf{x}}_i$ to satisfy (\ref{eq:target}), especially when $\epsilon$ is small. We evaluate the performance of the jamming module $g$ by measuring the \emph{accuracy} of the target model on the adversarial examples. A lower accuracy of the target model under adversarial attack indicates a better performance of the jamming model $g$, i.e., the more $\mathbf{x}_i$ satisfy (\ref{eq:target}), the better the performance of $g$ is.

    Next we describe how we construct the jamming module $g$. We first introduce FGSM, one of the most well-known adversarial example generators, and then our proposed UFGSM, which extends FGSM to unsupervised scenarios.

    \subsection{FGSM}

        FGSM was proposed by {Goodfellow et al. (2014)}~\cite{FGSM} and soon became a benchmark attack approach. Let $f$ be the target deep learning model, $\boldsymbol{\theta}$ be its parameters, and $J$ be the loss function in training $f$. The main idea of FGSM is to find an optimal max-norm perturbation $\eta$ constrained by $\alpha$ to maximize $J$. The perturbation can be calculated as:
        \begin{align}
            \eta = \alpha \cdot \mbox{sign}(\nabla_{\mathbf{x}_i} J(\boldsymbol{\theta},\mathbf{x}_i, y_i)) \label{eq:FGSM}
        \end{align}
        And hence the jamming module $g$ can be written as:
        \begin{align}
            g(\mathbf{x}_i) = \mathbf{x}_i + \alpha \cdot \mbox{sign}(\nabla_{\mathbf{x}_i}J(\boldsymbol{\theta},\mathbf{x}_i, y_i)) \label{eq:GeneratorFGSMA}
        \end{align}
        The requirement in (\ref{eq:limitation}) holds as long as $\alpha \leq \epsilon$.

        Let $\alpha=\epsilon$ so that we can perturb $\mathbf{x}_i$ at the maximum extent. Then, $g$ can be re-expressed as:
        \begin{align}
            g(\mathbf{x}_i) = \mathbf{x}_i + \epsilon \cdot \mbox{sign}(\nabla_{\mathbf{x}_i} J(\boldsymbol{\theta},\mathbf{x}_i, y_i)) \label{eq:GeneratorFGSM}
        \end{align}

        FGSM is an effective attack approach since it only requires calculating the gradients once instead of multiple times such as in the basic iterative method~\cite{BIM}.

    \subsection{White-Box Attack}

        In a white-box attack, the attacker knows the architecture and parameters $\boldsymbol{\theta}$ of the target model $f$. It may represent the scenario that a BCI system designer wants to evaluate the worst case performance of the system under attack, since usually white-box attacks cause more damages to the classifier than gray-box or black-box attacks. The designer then uses all information he/she knows about the classifier to attack it. It may also represent scenarios that the target model of a BCI system is somehow leaked/hacked, or the target model is publicly available.

        Knowing the architecture and parameters $\boldsymbol{\theta}$ of the target model $f$ is not enough for FGSM, because it needs to know the true label $y_i$ of $\mathbf{x}_i$ in order to generate the adversarial perturbation. Next we propose UFGSM, an unsupervised FGSM, to cope with this problem.

        UFGSM replaces the label $y_i$ by $y_i'=f(\mathbf{x}_i)$, i.e., the estimated label from the target model. Consequently, $g$ in UFGSM can be rewritten as:
        \begin{align}
            g(\mathbf{x}_i) = \mathbf{x}_i + \epsilon \cdot \mbox{sign}(\nabla_{\mathbf{x}_i}J(\boldsymbol{\theta},\mathbf{x}_i, y'_i)) \label{eq:GeneratorUFGSM}
        \end{align}
        $y_i'$ approaches $y_i$ when the accuracy of $f$ is high, and hence the performance of UFGSM approaches FGSM. However, as will be demonstrated in the next section, our proposed UFGSM is still effective even when $y_i'$ is quite different from $y_i$.

        The pseudocode of our proposed UFGSM for white-box attacks is shown in Algorithm~\ref{alg:white}.

        \begin{algorithm}
        \DontPrintSemicolon
        \KwIn{$f$, the target model; $\boldsymbol{\theta}$, the parameters of $f$; $J$, loss function of the target model; $\epsilon$, the upper bound of perturbation; $\mathbf{x}_i$, a normal EEG epoch.}
        \KwOut{$\widetilde{\mathbf{x}}_i$, an adversarial EEG epoch.\\
                    ~\\}
        Calculate $y_i'=f(\mathbf{x}_i)$;\\
        Calculate $\widetilde{\mathbf{x}}_i = \mathbf{x}_i + \epsilon \cdot \mbox{sign}(\nabla_{\mathbf{x}_i}J(\boldsymbol{\theta},\mathbf{x}_i, y_i'))$.\\
        ~\\
        \KwRet $\widetilde{\mathbf{x}}_i$
        \caption{Our proposed UFGSM for white-box attacks. }\label{alg:white}
        \end{algorithm}

    \subsection{Gray-Box Attack}

        UFGSM does not need the true labels when generating adversarial examples, but it assumes that we know the parameters of the target model $f$, which is still challenging in practice. This requirement can be eliminated by utilizing the transferability of adversarial examples, which is the basis of both gray-box and black-box attacks.

        The transferability of adversarial examples was first observed by {Szegedy et al. (2013)}~\cite{AdvExamSzegedy}, which may be the most dangerous property of adversarial examples. It denotes an intriguing phenomenon that adversarial examples generated by one deep learning model can also, with high probability, fool another model even the two models are different. This property has been used to attack deep learning models, e.g., {Papernot et al. (2016)}~\cite{BlackBoxAttack} attacked deep learning systems by training a substitute model with only queried information.

        To better secure deep learning systems, a lot of studies have been done to understand the transferability of adversarial examples. {Papernot et al. (2016)}~\cite{AdvTransferPapernot2016}, {Liu et al. (2016)}~\cite{AdvTransferLiu2016} and {Tramer et al. (2017)}~\cite{AdvTransferSpace} all attributed this property to some kind of similarity between the models. However, {Wu et al. (2018)}~\cite{PKTransAdv} questioned these explanations since they found that the transferability of adversarial examples is not symmetric, which does not satisfy the definition of similarity.

        Although more theoretical research is needed to understand both the adversarial example and its transferability property, this does not hinder us from using them in gray-box attack. Assume we have access to the training dataset used to construct the target model $f$, e.g., we know that $f$ was trained using some public databases. The basic idea of gray-box attack is to train our own model $f'$ to replace the target model $f$ in UFGSM, so that we can get rid of the dependency on the target model $f$.

        The pseudocode of UFGSM for gray-box attacks is shown in Algorithm~\ref{alg:gray}.

        \begin{algorithm}
        \DontPrintSemicolon
        \KwIn{$D$, training data of the target model; $J$, loss function of the substitute model; $\epsilon$, the upper bound of perturbation; $\mathbf{x}_i$, a normal EEG epoch.}
        \KwOut{$\widetilde{\mathbf{x}}_i$, an adversarial EEG epoch.\\
                    ~\\}
        Train a substitute model $f'$ from $D$, using loss function $J$; \\
        Calculate $y_i'=f'(\mathbf{x}_i)$; \\
        Calculate $\widetilde{\mathbf{x}}_i = \mathbf{x}_i + \epsilon \cdot \mbox{sign}(\nabla_{\mathbf{x}_i}J(\boldsymbol{\theta},\mathbf{x}_i, y_i'))$, where $\boldsymbol{\theta}$ encodes the parameters of $f'$.\\
        ~\\
        \KwRet $\widetilde{\mathbf{x}}_i$
        \caption{Our proposed UFGSM for gray-box attacks. }\label{alg:gray}
        \end{algorithm}

    \subsection{Black-Box Attack}

        Gray-box attack assumes the attacker has access to the training data of the target model, e.g., the target model is trained on some classic public datasets. An even more challenging situation is black-box attack, in which the attacker has access to neither the parameters of the target model nor its training data. Instead, the attacker can only interact with the target model and observe its output for an input. One example is to attack a commercial proprietary EEG-based BCI system. The attacker can buy such a system and observe its responses, but does not have access to the parameters or the training data of the target model.

        Papernot et al.~\cite{BlackBoxAttack} proposed an approach to perform black-box attacks. A similar idea is used in this paper. We record the inputs and outputs of the target model to train a substitute model $f'$, which is then used in UFGSM to generate adversarial examples, as shown in Algorithm~\ref{alg:black}. Note that the way we augment the training set is different from the original one in \cite{BlackBoxAttack}. In \cite{BlackBoxAttack}, the new training set was constructed by calculating the Jacobian matrix corresponding to the labels assigned to the inputs, whereas we use the loss computed from the inputs instead.

        \begin{algorithm}
        \DontPrintSemicolon
        \KwIn{$f$, the target model; $J$, loss function of the substitute model; $\lambda$, the parameter to control the step to generate the new training dataset; $N$, the maximum number of iterations; $\epsilon$, the upper bound of perturbation; $\mathbf{x}_i$, a normal EEG epoch.}
        \KwOut{$\widetilde{\mathbf{x}}_i$, an adversarial EEG epoch.\\
                    ~\\}
        Construct a set of unlabeled EEG epochs $S$;\\
        Pass $S$ through $f$ to generate a training dataset $D$;\\
        Train a substitute model $f'$ from $D$, using loss function $J$; \\
        \For{$n=1$ to $N$}
        {
            $\Delta S= \{\mathbf{x}+\lambda \cdot \mbox{sign}(\nabla_{\mathbf{x}}J(\boldsymbol{\theta},\mathbf{x}, y)):(\mathbf{x},y)\in D\}$, where $\boldsymbol{\theta}$ encodes the parameters of $f'$;\\
            $\Delta D=\{(\mathbf{x}_i,f(\mathbf{x}_i))\}_{\mathbf{x}_i \in \Delta S}$;\\
            $D \leftarrow D\bigcup\Delta D$;\\
            Train a substitute model $f'$ from $D$, using loss function $J$;
        }
        ~ \\
        Calculate $y_i'=f'(\mathbf{x}_i)$; \\
        Calculate $\widetilde{\mathbf{x}}_i = \mathbf{x}_i + \epsilon \cdot \mbox{sign}(\nabla_{\mathbf{x}_i}J(\boldsymbol{\theta},\mathbf{x}_i, y_i'))$, where $\boldsymbol{\theta}$ encodes the parameters of $f'$.\\
        ~\\
        \KwRet $\widetilde{\mathbf{x}}_i$
        \caption{Our proposed UFGSM for black-box attacks. }\label{alg:black}
        \end{algorithm}

\section{Experiments and Results} \label{sect:experiments}

    This section validates the performances of the three attack strategies. Three EEG datasets and three CNN models were used.

    \subsection{The Three EEG Datasets} \label{sect:datasets}

         The following three EEG datasets were used in our experiments:

        \subsubsection{P300 evoked potentials (P300)}
            The P300 dataset for binary-classification, collected from four disabled subjects and four healthy ones, was first introduced in \cite{EPFLP300}. In the experiment, a subjects faced a laptop on which six images were flashed randomly to elicit P300 responses. The goal was to classify whether the image is target or non-target. The EEG data were recorded from 32 channels at 2048Hz. We bandpass filtered the data to 1-40Hz and downsampled them to 256Hz. Then we extracted EEG epochs in [0,1]s after each image onset, normalized them using  $\frac{x-\footnotesize{\mbox{mean}}(x)}{10}$, and truncated the resulting values into [-5, 5], as the input to the CNN classifiers. Each subject had about 3,300 epochs.

        \subsubsection{Feedback error-related negativity (ERN)}
            The ERN dataset \cite{ERN} was used in a Kaggle competition\footnote{https://www.kaggle.com/c/inria-bci-challenge} for two-class classification. It was collected from 26 subjects and partitioned into training set (16 subjects) and test set (10 subjects). We only used the 16 subjects in the training set as we do not have access to the test set. The 56-channel EEG data had been downsampled to 200Hz. We bandpass filtered them to 1-40Hz, extracted EEG epochs between [0,1.3]s, and standardized them using $z$-score normalization. Each subject had 340 epochs.

        \subsubsection{Motor imagery (MI)}
            The MI dataset is Dataset 2A\footnote{http://www.bbci.de/competition/iv/} in BCI Competition IV~\cite{MI4C}. It was collected from nine subjects and consisted of four classes (imagined movements of the left hand, right hand, feet, and tongue). The 22-channel EEG signals were sampled at 128Hz. As in \cite{EEGNet}, we bandpass filtered the data to 4-40Hz, and standardized them using an exponential moving average window with a decay factor of 0.999. Each subject had 576 epochs, 144 in each class.

    \subsection{The Three CNN Models}

        The following three different CNN models were used in our experiments:

        \subsubsection{EEGNet}
            EEGNet \cite{EEGNet} is a compact CNN architecture with only about 1000 parameters (the number may change slightly according to the nature of the task). EEGNet contains an input block, two convolutional blocks and a classification block. To reduce the number of model parameters, it replaces the traditional convolution operation with a depthwise separable convolution, which is the most important block in Xception~\cite{Xception}.

        \subsubsection{DeepCNN}
            Compared with EEGNet, DeepCNN\cite{MNE} is deeper and hence has much more parameters. It consists of four convolutional blocks and a classification block. The first convolutional block is specially designed to handle EEG inputs and the other three are standard ones.

        \subsubsection{ShallowCNN}
            ShallowCNN \cite{MNE} is a shallow version of DeepCNN, inspired by filter bank common spatial patterns~\cite{FBCSP}. Its first block is similar to the first convolutional block of DeepCNN, but with a larger kernel, a different activation function, and a different pooling approach.

    \subsection{Training Procedure and Performance Measures}

    The first two datasets have high class imbalance. To accommodate this, in training we applied a weight to each class, which was the inverse of its number of examples in the training set. We used the Adam optimizer~\cite{Adam}, and cross entropy as our loss function. Early stopping was used to reduce overfitting.

    The test set still had class imbalance, which resembled the practical application scenario. We employed two metrics to evaluate the test performance:
     \begin{enumerate}
      \item \textit{Raw classification accuracy} (RCA), which is the ratio of the total number of correctly classified test examples to the total number of test examples.
      \item \textit{Balanced classification accuracy} (BCA) \cite{drwuTHMS2017}, which is the average of the individual RCAs of different classes.
      \end{enumerate}

    \subsection{Baseline Performances on Clean EEG Data}

        We first evaluated the baseline performances of the three CNN models.

        \textbf{Within-subject experiments} -- For each individual subject, epochs were shuffled and divided into 80\% training and 20\% test. We further randomly sampled 25\% epochs from the training set as our validation set in early stopping. We calculated the mean RCAs and BCAs from all subjects as the performance measures.

        \textbf{Cross-subject experiments} -- For each dataset, leave-one-subject-out cross-validation was performed, and the mean RCAs and BCAs were calculated. Epochs from all subjects in the training set were mixed, shuffled, and divided into 75\% training and 25\% validation.

\begin{figure}[htbp]\centering
            \subfigure[]{\label{fig:result_clean_RCA}   \includegraphics[width=.9\linewidth,clip]{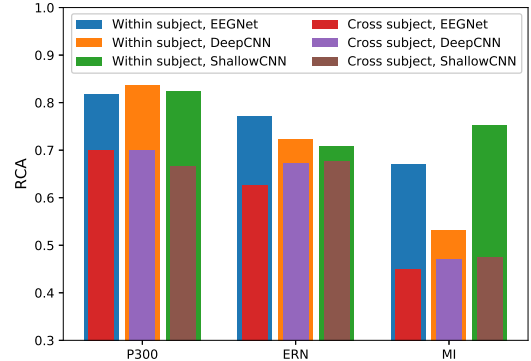}}
            \subfigure[]{\label{fig:result_clean_BCA}    \includegraphics[width=.9\linewidth,clip]{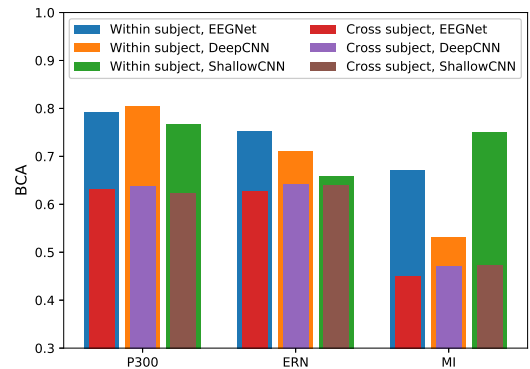}}
        \caption{Baseline classification accuracies of the three CNN classifiers on different datasets. (a) RCAs; (b) BCAs.} \label{fig:clean}
        \end{figure}

        The baseline results are shown in Fig.~\ref{fig:clean} and Table~\ref{tab:results}, and the corresponding models were regarded as our target models. Note that the RCAs and BCAs on the MI dataset were considerably lower, because MI was 4-class classification, whereas P300 and ERN were 2-class classification. For all datasets and all classifiers, RCAs and BCAs of within-class experiments were higher than their counterparts in cross-subject experiments, which is reasonable, because individual differences cause inconsistency among examples from different subjects.

        \begin{table*}[htbp] \center
        \caption{RCAs/BCAs of different CNN classifiers in white-box and gray-box attacks on the three datasets ($\epsilon=0.1/0.1/0.05$ for P300/ERN/MI). }   \label{tab:results}
        \begin{tabular}{c|c|c|c|c|c|ccc}
        \toprule
        \multirow{2}{*}{Experimental Setup} &\multirow{2}{*}{Dataset}   &\multirow{2}{*}{Target Model $f$}    &\multicolumn{2}{c|}{Baselines}  &White-Box  &\multicolumn{3}{c}{Substitute Model $f'$ in Gray-Box Attack} \\ \cline{4-5}\cline{7-9}
             &  &   &Clean Data  &Noisy Data  &Attack  &EEGNet         &DeepCNN        &ShallowCNN\\
        \midrule

        \multirow{9}{*}{Within-Subject} &\multirow{3}{*}{P300}  &EEGNet     &$.8168/.7915$  &$.8156/.7896$  &$.1947/.2204$  &$.2342/.2486$  &$.2832/.3017$  &$.6753/.6538$\\
                                        &                       &DeepCNN    &$.8371/.8049$  &$.8350/.8006$  &$.2047/.2276$  &$.3865/.3717$  &$.3065/.3118$ &$.7110/.6730$\\
                                        &                       &ShallowCNN &$.8243/.7668$  &$.8264/.7623$  &$.1868/.2420$  &$.6764/.6178$  &$.6388/.5889$ &$.3238/.3335$\\ \cline{2-9}
                                        &\multirow{3}{*}{ERN}   &EEGNet     &$.7702/.7513$  &$.7693/.7435$  &$.2353/.2519$  &$.4283/.4070$  &$.5377/.5268$ &$.6994/.6697$\\
                                        &                       &DeepCNN    &$.7224/.7100$  &$.7233/.7113$  &$.3539/.3549$  &$.5864/.5689$  &$.5294/.4983$ &$.6719/.6483$\\
                                        &                       &ShallowCNN &$.7077/.6589$  &$.7068/.6577$  &$.3162/.3450$  &$.6811/.6279$  &$.6535/.6062$ &$.4770/.4855$\\ \cline{2-9}
                                        &\multirow{3}{*}{MI}    &EEGNet     &$.6705/.6698$  &$.6552/.6551$  &$.1561/.1570$  &$.2337/.2312$  &$.4684/.4759$ &$.3554/.3593$\\
                                        &                       &DeepCNN    &$.5316/.5316$  &$.5250/.5248$  &$.2586/.2609$  &$.5067/.5067$  &$.3736/.3803$ &$.3554/.3589$\\
                                        &                       &ShallowCNN &$.7519/.7505$  &$.7213/.7212$  &$.1705/.1728$  &$.6130/.6179$  &$.5575/.5636$ &$.2289/.2296$\\
        \midrule
        \multirow{9}{*}{Cross-Subject}  &\multirow{3}{*}{P300}  &EEGNet     &$.6985/.6306$  &$.6978/.6295$  &$.3085/.3786$  &$.3635/.4092$
                                        &$.4666/.4729$ &$.5351/.5308$\\
                                        &                       &DeepCNN    &$.6992/.6366$  &$.6982/.6345$  &$.3095/.3666$  &$.4096/.4500$   &$.3739/.4115$  &$.4711/.4882$\\
                                        &                       &ShallowCNN &$.6659/.6225$  &$.6660/.6230$  &$.3346/.3783$  &$.4429/.4799$   &$.4439/.4694$  &$.3509/.4013$\\ \cline{2-9}
                                        &\multirow{3}{*}{ERN}   &EEGNet     &$.6250/.6266$  &$.6272/.6309$  &$.3904/.3823$  &$.3561/.3917$   &$.3263/.3533$  &$.5241/.5455$\\
                                        &                       &DeepCNN    &$.6719/.6404$  &$.6792/.6434$  &$.3281/.3595$  &$.3450/.3897$   &$.3254/.3464$  &$.5379/.5186$\\
                                        &                       &ShallowCNN &$.6754/.6394$  &$.6783/.6391$  &$.3246/.3606$  &$.5555/.5533$   &$.5438/.5125$  &$.3379/.3702$\\ \cline{2-9}
                                        &\multirow{3}{*}{MI}    &EEGNet     &$.4500/.4500$  &$.4460/.4460$  &$.2369/.2369$  &$.2531/.2531$   &$.2834/.2834$  &$.2785/.2785$\\
                                        &                       &DeepCNN    &$.4695/.4695$  &$.4655/.4655$  &$.2550/.2550$  &$.3536/.3536$   &$.2865/.2865$  &$.2948/.2948$\\
                                        &                       &ShallowCNN &$.4734/.4734$  &$.4660/.4660$  &$.2610/.2610$  &$.3520/.3520$   &$.3009/.3009$  &$.2658/.2658$\\
        \bottomrule
        \end{tabular}
        \end{table*}

       \subsection{Baseline Performances under Random Noise}

        Before attacking these three target models using our proposed approaches, we corrupted the clean EEG data with random noise $\eta_0$ to check if that can significantly deteriorate the classification performances. If so, then we do not need to use a sophisticated approach to construct the adversarial examples: just adding random noise is enough.

        The random noise was designed to be:
        \begin{align}
            \eta_0 = \epsilon \cdot \mbox{sign}\left(\mathcal{N}(0,1)\right) \label{eq:RandomNoise}
        \end{align}
        i.e., $\eta_0$ is either $-\epsilon$ or $\epsilon$, so that its amplitude resembles that of the adversarial perturbations in (\ref{eq:limitation}). Although the EEG in all three datasets had similar standard deviations, empirically we found that the CNN classifiers trained on the MI dataset were more sensitive to noise than those on P300 or ERN. So, we set $\epsilon=0.1$ for P300 and ERN, and $\epsilon=0.05$ for MI in the experiments.

        The results are shown in Table~\ref{tab:results}. The classification accuracies on the noisy EEG data were comparable with, and sometimes even slightly better than, those on the clean EEG data, suggesting that adding random noise did not have a significant influence on the target models. In other words, adversarial perturbations cannot be implemented by simple random noise; instead, they must be deliberately designed.

    \subsection{White-Box Attack Performance}

        In a white-box attack, we know the target model exactly, including its architecture and parameters. Then, we can use UFGSM in Algorithm~1 to attack the target model. The results are shown in Table~\ref{tab:results}. Clearly, there were significant performance deteriorations in all cases, and in most cases the classification accuracies after attack were even lower than random guess. Interestingly, though UFGSM is based on the assumption that the target model should have high accuracy so that we can replace the true class labels with the predicted ones, Table~\ref{tab:results} shows that significant damages could also be made even when the target model has low accuracy, e.g., on the MI dataset.

\begin{figure}[htbp]   \centering
        \includegraphics[width=\linewidth,clip]{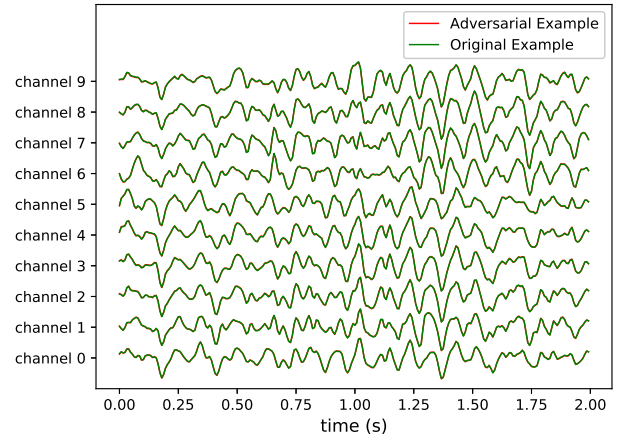}
        \caption{An example of the EEG epoch before and after adversarial perturbation (MI dataset). $\epsilon=0.05$.} \label{fig:examples}
        \end{figure}

        An example of the EEG epoch before and after adversarial perturbation is shown in Fig.~\ref{fig:examples}. The perturbation was so small that it is barely visible, and hence difficult to detect.

        In summary, our results showed that the three CNN classifiers can all be easily fooled with tiny adversarial perturbations generated by the proposed UFGSM in Algorithm~1, when the attacker has full knowledge of the target model.

\begin{figure}[htbp]\centering
            \subfigure[]{\label{fig:E_P300}   \includegraphics[width=.8\linewidth,clip]{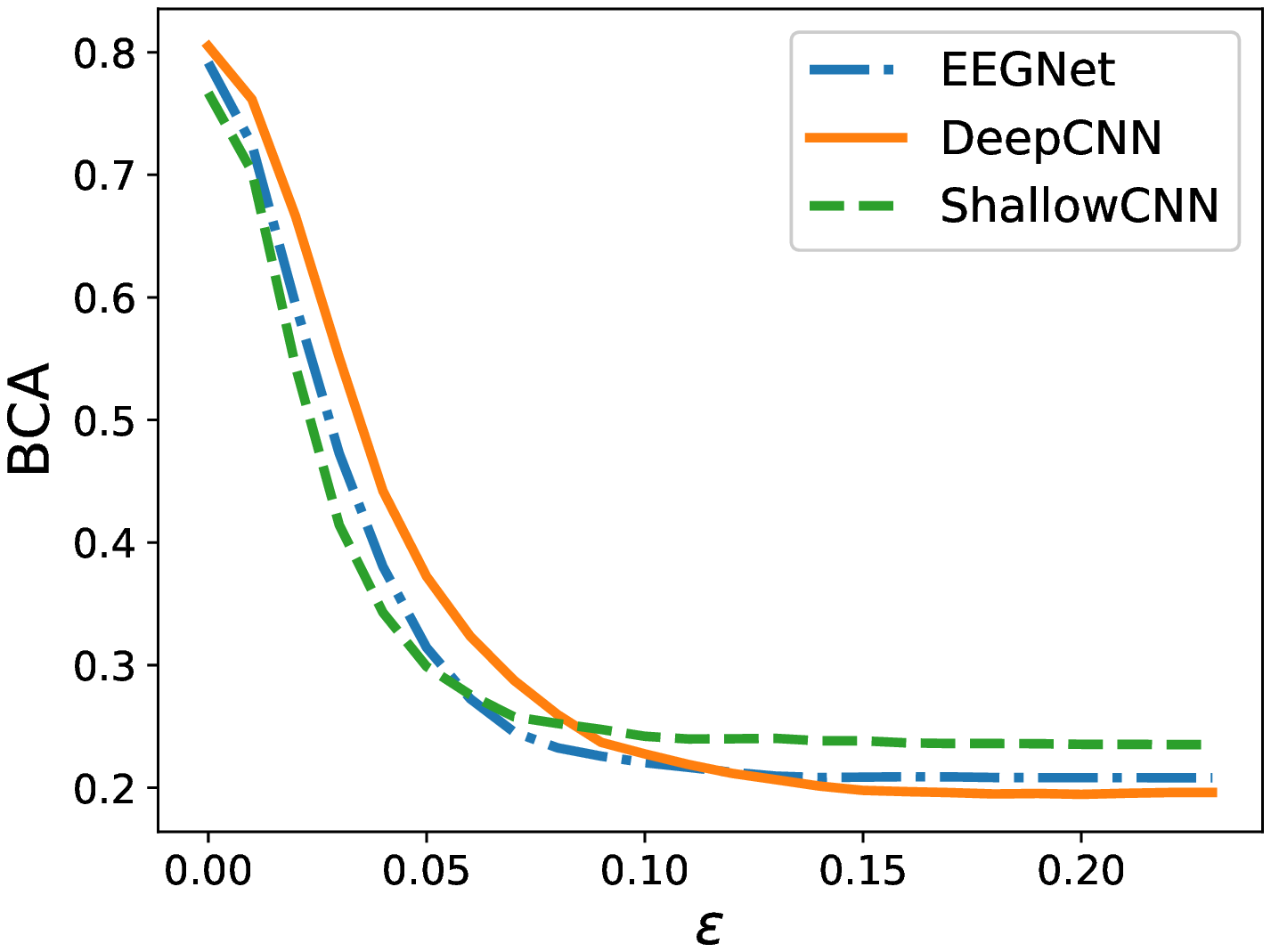}}
            \subfigure[]{\label{fig:E_ERN}    \includegraphics[width=.8\linewidth,clip]{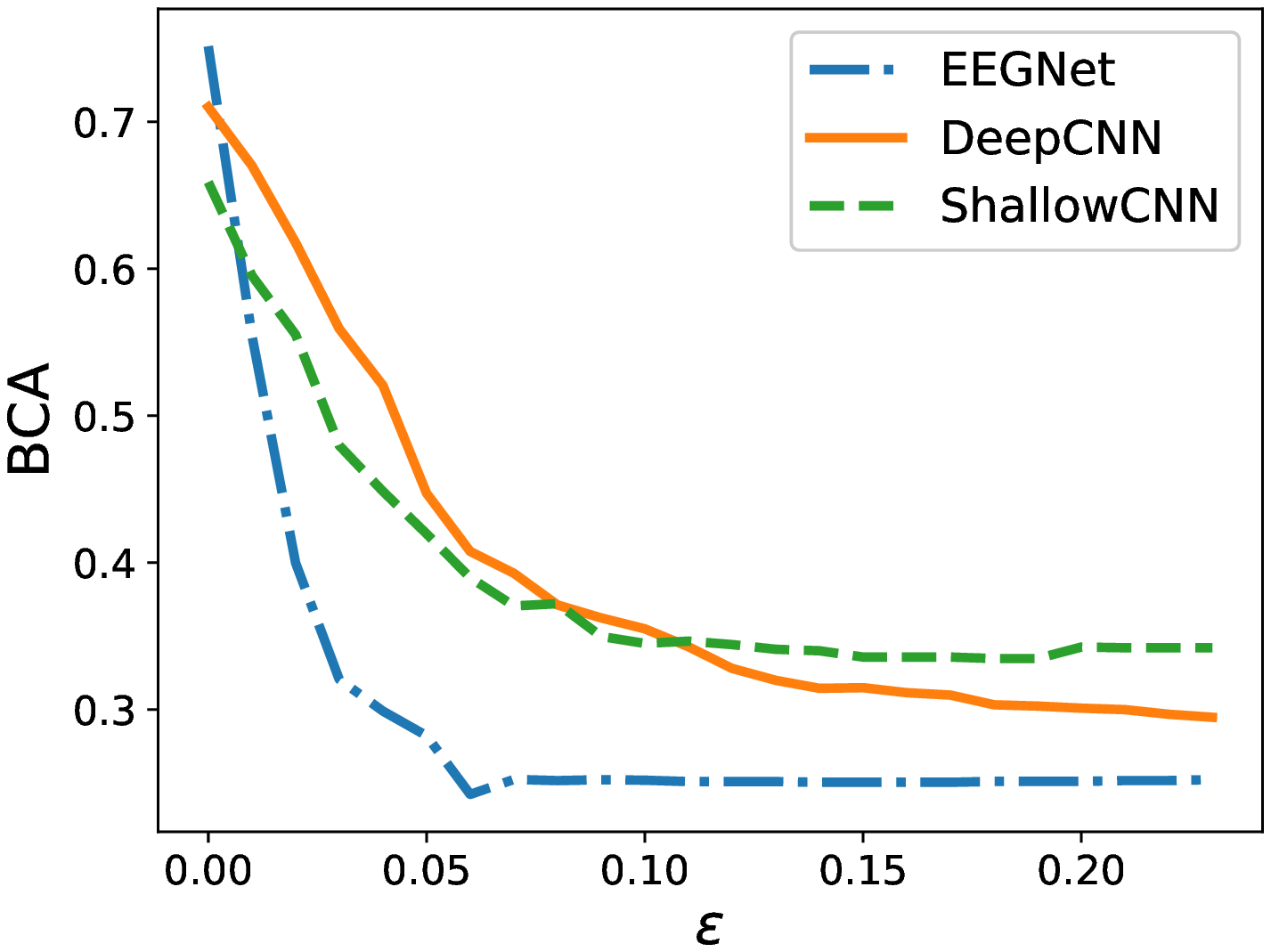}}
            \subfigure[]{\label{fig:E_MI}     \includegraphics[width=.8\linewidth,clip]{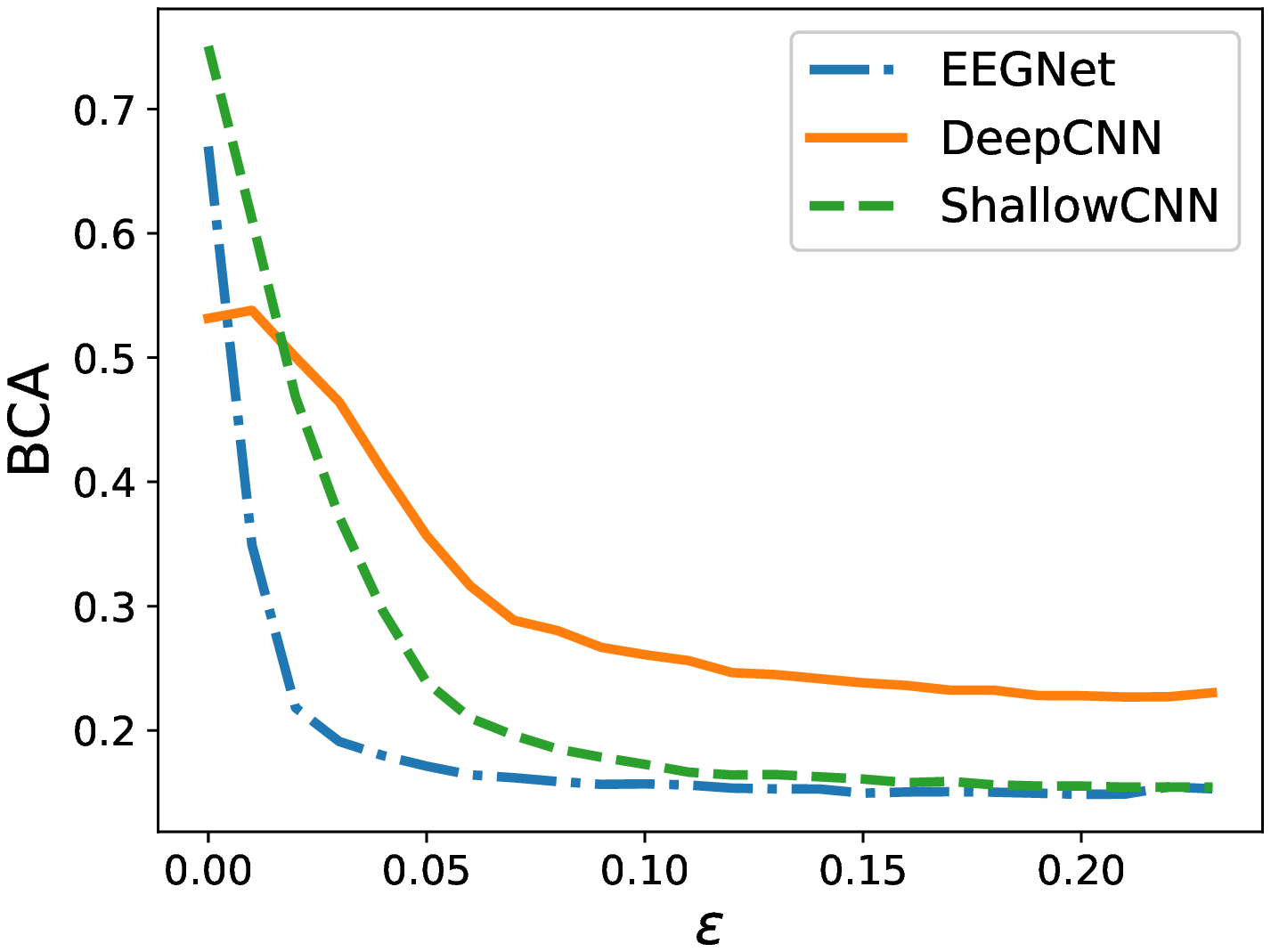}}
        \caption{BCAs of the target model after within-subject white-box attack, with respect to different $\epsilon$. a) P300 dataset; b) ERN dataset; and, c) MI dataset.} \label{fig:epsilon}
        \end{figure}

        $\epsilon=0.1$ for P300/ERN and $\epsilon=0.05$ for MI were used in the above experiments. Since $\epsilon$ is an important parameter in Algorithm~1, we also evaluated the performance of white-box attack with respect to different values of $\epsilon$. The results are shown in Fig.~\ref{fig:epsilon}. In all cases, the post-attack accuracy first decreased rapidly as $\epsilon$ increased, and then converged to a low value.

    \subsection{Gray-Box Attack Performance}

        Gray-box attack considers a more practical scenario than white-box attack: instead of knowing the architecture and parameters of the target model, here we only know its training data. In gray-box attack, we train a substitute model $f'$ on the same training data, and use it in Algorithm~2 to generate adversarial examples. This subsection verifies the effectiveness of gray-box attack. Again, we set $\epsilon=0.1/0.1/0.05$ for P300/ERN/MI, respectively.

        Assume the target model is EEGNet, but the attacker does not know. The attacker randomly picks a model architecture, e.g., DeepCNN, and trains it using the known training data. This model then becomes the substitute model $f'$ and is used in Algorithm~2 to generate adversarial examples. When different target models and different substitute models are used, the attacking performances are shown in the last part of Table~\ref{tab:results} for the three datasets. We can observe that:
        \begin{enumerate}
        \item The RCAs and BCAs after gray-box attacks were lower than the corresponding baseline accuracies, suggesting the effectiveness of the proposed gray-box attack approach.
        \item The RCAs and BCAs after gray-box attacks were generally higher than the corresponding accuracies after white-box attacks, especially when the attacker did not guess the architecture of the target model right, suggesting that knowing more target model information (white-box attack) can lead to more effective attacks.
        \item In gray-box attacks, when the attacker guessed the right architecture of the target model, the attack performance was generally better than when he/she guessed the wrong architecture.
        \end{enumerate}

    \subsection{Black-Box Attack Performance}

        This subsection considers a hasher but most practical scenario: the attacker knows neither the parameters nor the training set of the target model.

        To simulate such a scenario, we partitioned \textit{8/16/9} subjects in the P300/ERN/MI dataset into two groups: \textit{7/14/7} subjects in Group~A, and the remaining \textit{1/2/2} in Group~B. We assume that the CNN classifier in the EEG-based BCI system was trained on Group~A. The attacker, who belongs to Group~B, bought such a system and would like to collect some data from himself/herself, train a substitute model $f'$ using Algorithm~\ref{alg:black}, and then attack the CNN classifier. It's important to note that we used 80\% epochs in Group~A for training the target model $f$ (among which 75\% were used for tuning the parameters, and 25\% for validation in early stopping), and the remaining 20\% epochs in Group~A for testing $f$ and $f'$. Before training on the P300 and ERN datasets, to balance the classes of our initial dataset, we randomly downsampled the majority class according to the labels that the target model predicted at the first time.

        $\lambda=0.5$ and $N=2$ (Algorithm~3) were used in our experiments. We only performed black-box attack on the mixed epochs from all the subjects in the training set, since it was too time-consuming to train cross-subject models or within-subject model for each subject. The baseline and black-box attack results are shown in Table~\ref{tab:black}. Note that the baseline results were slightly different from those in Table~\ref{tab:results}, because here we only used a subset of the subjects to train the baseline models, whereas previously we used all subjects. After black-box attack, the RCAs and BCAs of all target models decreased, suggesting that our proposed black-box attack strategy was effective. Generally, when the substitute model $f'$ had the same structure as the target model $f$, e.g., both were EEGNet, the attack was most effective. This is intuitive.

          \begin{table*}[htbp] \center
        \caption{Mixed-subject RCAs/BCAs before and after black-box attack on the three datasets ($\epsilon=0.1/0.1/0.05$ for P300/ERN/MI).}  \setlength{\tabcolsep}{1.3mm}  \label{tab:black}
        \begin{tabular}{c|c|cc|ccc}
        \toprule
        \multirow{2}{*}{Dataset}&\multirow{2}{*}{Target Model $f$}  &\multicolumn{2}{c}{Baselines}
        &\multicolumn{3}{|c}{Substitute Model $f'$}       \\  \cline{3-4} \cline{5-7}
                    &            &Clean         &Noisy          &EEGNet         &DeepCNN        &ShallowCNN\\
        \midrule
                    &EEGNet      &$.7570/.7179$ &$.7531/.7156$  &$.3955/.4188$  &$.5244/.5506$  &$.5212/.5559$    \\
        P300        &DeepCNN     &$.7713/.7404$ &$.7747/.7430$  &$.3957/.4081$  &$.3589/.4297$  &$.4617/.4806$    \\
                    &ShallowCNN  &$.7336/.7163$ &$.7375/.7186$  &$.6315/.6189$  &$.5113/.5505$  &$.4118/.4301$    \\
                    \midrule
                    &EEGNet      &$.7665/.7614$ &$.7687/.7640$  &$.3113/.3288$  &$.4893/.5024$  &$.7207/.7006$    \\
        ERN         &DeepCNN     &$.7719/.7455$ &$.7740/.7519$  &$.3134/.3644$  &$.3049/.3963$  &$.6972/.7113$    \\
                    &ShallowCNN  &$.7495/.7399$ &$.7367/.7238$  &$.4478/.3860$  &$.3977/.3906$  &$.4691/.5715$    \\
                    \midrule
                    &EEGNet      &$.5603/.5565$ &$.5345/.5324$  &$.2352/.2343$  &$.2586/.2571$  &$.3005/.2988$    \\
        MI          &DeepCNN     &$.5222/.5189$ &$.5135/.5108$  &$.4446/.4446$  &$.4483/.4475$  &$.4384/.4353$    \\
                    &ShallowCNN  &$.6201/.6201$ &$.6133/.6128$  &$.5394/.5411$  &$.5062/.5045$  &$.4433/.4466$    \\
        \bottomrule
        \end{tabular}
        \end{table*}

    \subsection{Characteristics of the Perturbations}

        To better understand the characteristics of adversarial perturbations, this subsection studies the signal-to-noise ratio (SNR) of the adversarial examples, and the spectrogram of the perturbations.

        We had no clue of the SNR of the normal epochs, so we had to assume that they contained very little noise. To compute the SNR of the noisy epochs [the random noise was generated using (\ref{eq:RandomNoise})], we treated the normal epochs as the clean signals, and the noise in (\ref{eq:RandomNoise}) as the noise. To compute the SNR of the adversarial examples, we treated the normal epochs as the clean signals, and the adversarial perturbations as noise. The SNRs are shown in Table~\ref{tab:SNR}. In all three datasets, the SNRs of the noisy examples were roughly the same as those in the adversarial examples, which is intuitive, since they were controlled by the same parameter $\epsilon$ in our experiments. With the same amount of noise, the deliberately generated perturbations can significantly degrade the performances of the CNN classifiers, whereas random noise cannot, suggesting again the effectiveness of our proposed algorithms.

        \begin{table}[htbp] \center
        \caption{SNRs (dB) of noisy examples (normal examples plus random noise) and adversarial examples. $\epsilon=0.1/0.1/0.05$ for P300/ERN/MI.}   \label{tab:SNR}
            \begin{tabular}{c|c|ccc}
            \toprule
             \multirow{2}{*}{Dataset} &\multirow{2}{*}{Noisy examples}  &\multicolumn{3}{|c}{Adversarial examples}\\ \cline{3-5}
                    &               &EEGNet     &DeepCNN    &ShallowCNN \\ \midrule
            P300    &20.43          &20.43      &20.50      &20.53      \\
            ERN     &20.26          &20.26      &20.39      &20.31      \\
            MI      &25.50          &25.50      &25.57      &25.60      \\
            \bottomrule
            \end{tabular}
        \end{table}

\begin{figure*}[htbp]\centering
            \subfigure[]{\label{fig:P300_EEGNet_Spec}   \includegraphics[width=.32\linewidth,clip]{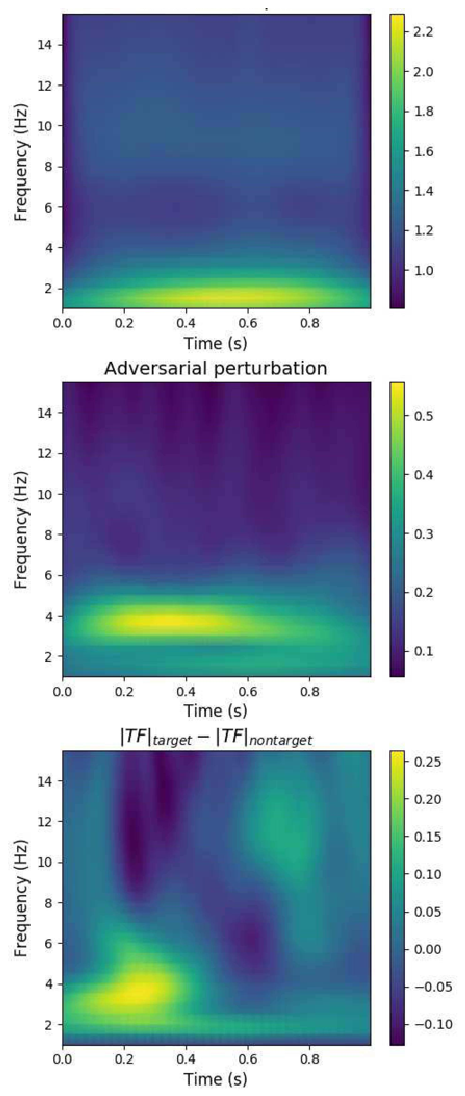}}
            \subfigure[]{\label{fig:P300_DeepCNN_Spec}    \includegraphics[width=.32\linewidth,clip]{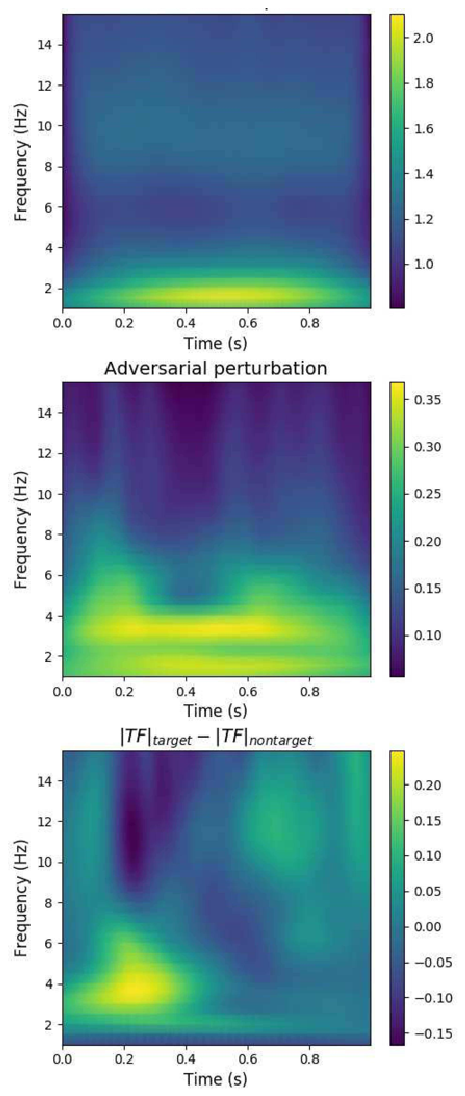}}
            \subfigure[]{\label{fig:P300_ShallowCNN_Spec}     \includegraphics[width=.32\linewidth,clip]{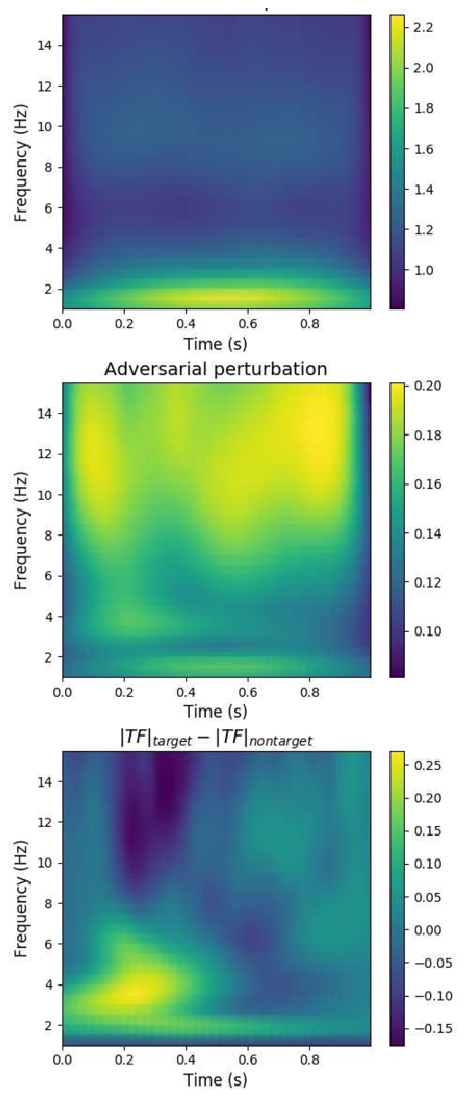}}
        \caption{Mean spectrogram of the normal examples (top row), mean spectrogram of the corresponding perturbations (middle row), and mean spectrogram difference between target and non-target normal examples (bottom row), in within-subject white-box attack on the P300 dataset. Note that the scales are different. The channel was randomly chosen. (a) EEGNet; (b) DeepCNN; and, (c) ShallowCNN.} \label{fig:Spec}
        \end{figure*}

        Next we analyze the spectrogram of the adversarial examples. Consider within-subject white-box attacks on the P300 dataset. For each classifier we partitioned the misclassified adversarial examples into two groups. Group~1 consisted of non-target examples whose adversarial examples were classified as targets, and Group~2 consisted of target examples whose adversarial examples were classified as non-targets. We then computed the spectrograms of all such examples using wavelet decomposition, the mean spectrogram of each group, and the difference of the two group means. The results are shown in the first and third row of Fig.~\ref{fig:Spec} for the three classifiers. They were very similar to each other, in terms of their patterns and ranges. We could observe a clear peak around 0.2s for all three classifiers, maybe corresponding to the onset of P300 responses.

        We then computed the mean spectrogram of the adversarial perturbations. The results are shown in the second row of Fig.~\ref{fig:Spec} (note that their scales were much smaller than those in the first row). The patterns and ranges are now noticeably different. For EEGNet, the energy of the perturbations was concentrated in a small region, i.e., $[0, 0.8]$s and $[3, 5]$Hz, whereas that for DeepCNN was a little more scattered, and that for ShallowCNN was almost uniformly distributed in the entire time-frequency domain. These results suggest that the perturbations from the three CNN classifiers had dramatically different shapes, maybe specific to the characteristics of the classifiers. They also explain the within-subject gray-box attack results on the P300 dataset in Table~\ref{tab:results}: the perturbations generated from EEGNet and DeepCNN were similar, and hence EEGNet (DeepCNN) as a substitute model could effectively attack DeepCNN (EEGNet). However, the perturbations generated from EEGNet/DeepCNN and ShallowCNN were dramatically different, and hence EEGNet/DeepCNN were less effective in attacking ShallowCNN, and vice versa.

\subsection{Additional Attacks} \label{sect:Discussion}

         To increase the robustness of a P300-based BCI system, a CNN classifier may be applied to the synchronized average of multiple epochs, instead of a single epoch. It's interesting to know if this averaging strategy can help defend adversarial attacks.

         As mentioned in Section~\ref{sect:datasets}, the P300 dataset was collected from eight subjects. Each subject completed four recording sessions, and each session consisted of six runs, one for each of the six images. The number of epochs of each run was about 120 as each image was flashed about 20 times. We constructed an averaged epoch as the average of 10 epochs from the same image. Thus, two averaged epochs were obtained from each image, and for each subject, $4\times6\times2=48$ averaged target (P300) epochs were obtained. Similarly, we obtained 248 averaged non-target epochs for each subject. These epochs were shuffled and divided into 80\% training and 20\% test for each subject.

\begin{figure}[htbp]\centering
            \subfigure[]{\label{fig:PSE}   \includegraphics[width=.2\linewidth,clip]{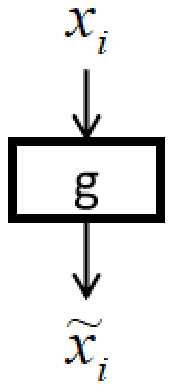}}\qquad
            \subfigure[]{\label{fig:AAE}   \includegraphics[width=.2\linewidth,clip]{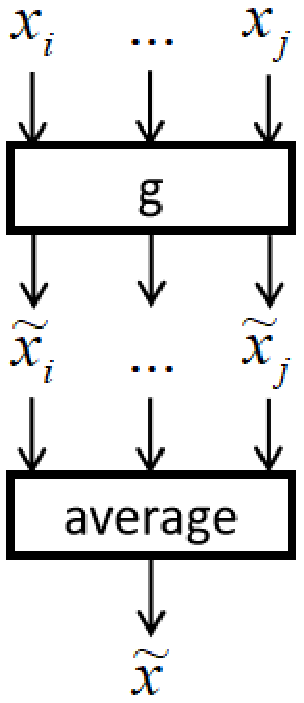}}\qquad
            \subfigure[]{\label{fig:PAE}   \includegraphics[width=.2\linewidth,clip]{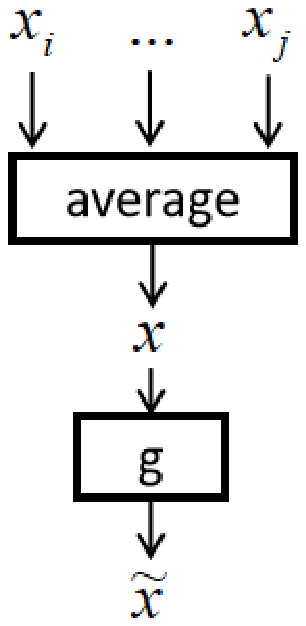}}
        \caption{Three approaches to generate adversarial examples on the synchronized-averaging epochs. (a) PSE (no average); (b) AAE (attack, then average); and, (c) PAE (average, then attack).} \label{fig:Average}
        \end{figure}

         We then compared the following three white-box attacks ($\epsilon=0.1$) in within-subject experiments:
         \begin{enumerate}
           \item \textit{Perturbation on each single epoch} (PSE), in which an adversarial example was generated for each single (un-averaged) epoch, as shown in Fig.~\ref{fig:PSE}. This was also the main attack considered before this subsection.
           \item \textit{Averaged adversarial examples} (AAE), in which an adversarial example was generated for each of the 10 single epochs, as in PSE, and then their average was taken, as shown in Fig.~\ref{fig:AAE}.
           \item \textit{Perturbation on the averaged epochs} (PAE), in which an adversarial example was generated directly on each averaged epoch, as shown in Fig.~\ref{fig:PAE}.
         \end{enumerate}

         To better explore the transferability of adversarial examples, we also tested a traditional approach\footnote{https://github.com/alexandrebarachant/bci-challenge-ner-2015}, xDAWN+RG, which won the Kaggle BCI competition in 2015 and was also tested in \cite{EEGNet}. xDAWN+RG used xDAWN spatial filtering~\cite{Rivet2009}, Riemannian geometry~\cite{Barachant2012} and ElasticNet to classify the epochs. We attacked it using PAE adversarial examples generated by different CNN models.

          The results are shown in Table~\ref{tab:multi_single}\footnote{The baseline single epoch attack results were different from those in Table~II, especially for ShallowCNN, because here the CNN classifiers were trained on the averaged epochs, and then applied to the single epochs, whereas in Table~II the CNN classifiers were trained directly on the single epochs.}. All four approaches had improved performance when trained and tested on the averaged epochs, suggesting the rationale to take the synchronized average (of course, this also has the side effect of reducing the speed of the BCI system). For each CNN model, all three attack approaches were effective, but the attack was most effective when the adversarial examples were generated on the averaged epochs (i.e., PAE). The second part of Table~\ref{tab:multi_single} shows that adversarial examples generated by CNN models could be used to attack xDAWN+RG, but not as effective as in attacking the CNN models. However, this does not mean that traditional machine learning approaches are example from adversarial attacks \cite{papernot2016transferability}. They may require different attack strategies.

        \begin{table*}[htbp] \center
        \caption{Within-subject RCAs/BCAs before and after white-box attacks on the averaged P300 epochs. $\epsilon=0.1$.}   \label{tab:multi_single}
            \begin{tabular}{c|cc|cc|ccc}
            \toprule
            \multirow{2}{*}{models} &\multicolumn{4}{c}{Baseline}   &\multicolumn{3}{|c}{Adversarial examples}\\ \cline{2-5} \cline{6-8}
                                    &Single             &Noisy          &Average        &Noisy          &PSE            &AAE            &PAE \\ \midrule
            EEGNet                  &.7824/.7184        &.7818/.7200    &.9569/.9677    &.9461/.9443    &.1272/.2441    &.2459/.3001    &.0431/.0323      \\
            DeepCNN                 &.7725/.7375        &.7712/.7383    &.9332/.9265    &.9224/.9202    &.2371/.3494    &.3169/.3329    &.0819/.0826      \\
            ShallowCNN              &.2477/.5291        &.2417/.5273    &.8685/.8393    &.8556/.8357    &.8491/.6147    &.5465/.5421    &.1315/.1607      \\ \midrule
                                    &                   &               &               &            &\multicolumn{2}{c}{Attacked by EEGNet}     &.7931/.6417 \\
            xDAWN+RG                &.8145/.5697        &.8000/.5654    &.9569/.8934    &.9289/.7775 &\multicolumn{2}{c}{Attacked by DeepCNN}    &.7177/.5501 \\
                                    &                   &               &               &            &\multicolumn{2}{c}{Attacked by ShallowCNN} &.8664/.6964 \\
            \bottomrule
            \end{tabular}
        \end{table*}

\begin{figure}[htbp]   \centering
        \includegraphics[width=\linewidth,clip]{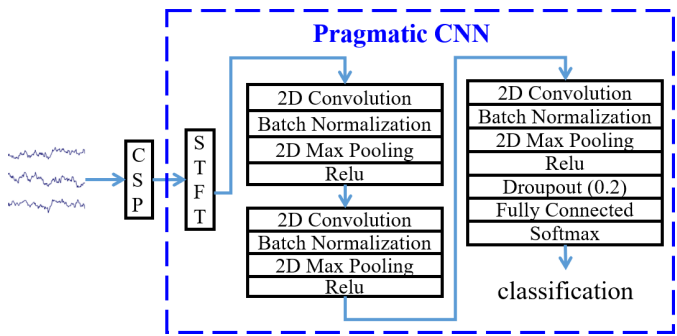}
        \caption{The CNN classification pipeline on the MI dataset, when EEG spectrogram is used as the input feature.} \label{fig:SpecCNN}
        \end{figure}

         In all previous experiments we used raw EEG signals as the input to the CNN models. However, the spectrograms are also frequently used in MI based BCIs. Next, we study whether our attack strategies can still work for the spectrogram input. The CNN classification pipeline is shown in Fig.~\ref{fig:SpecCNN}, where common spatial pattern (CSP) filtering \cite{Koles1990} was used to reduce the number of EEG channels from 22 to 8, short-time Fourier transform (STFT) was used to convert EEG signals into spectrograms, and PragmatricCNN \cite{Tayeb2019} was used as the classifier (EEGNet, DeepCNN and ShallowCNN cannot work on the spectrograms). Because both CSP and STFT are differentiable operations, we can compute the gradients over the whole pipeline to find the adversarial perturbations on the raw EEG time series.

         The within-subject white-box attack results on the MI dataset are shown in Table~\ref{tab:pragmatic_result}. Clearly, the attack was very successful, suggesting that simply extracting the spectrogram as the input features cannot effectively defend adversarial attacks.

        \begin{table}[htbp] \center
        \caption{Within-subject RCAs/BCAs before and after white-box attacks on PragmaticCNN for the MI dataset. $\epsilon=0.05$.}   \label{tab:pragmatic_result}
            \begin{tabular}{c|cc|c}
            \toprule
            Subject        &Clean Examples     &Noisy Examples     & Adversarial Examples\\ \midrule
            s1              &.8103/.8012        &.8103/.7958        &.1638/.1754\\
            s2              &.6552/.6422        &.6638/.6508        &.1810/.1898\\
            s3              &.8621/.8525        &.8017/.8006        &.1379/.1466\\
            s4              &.6810/.6781        &.6724/.6722        &.1983/.2023\\
            s5              &.5862/.5946        &.5690/.5727        &.2672/.2609\\
            s6              &.5259/.5272        &.5517/.5532        &.2241/.2250\\
            s7              &.8448/.8491        &.8103/.8194        &.1810/.1752\\
            s8              &.8448/.8328        &.7759/.7568        &.2414/.2535\\
            s9              &.8707/.8673        &.8448/.8411        &.2414/.2299\\ \midrule
            Average         &.7423/.7383        &.7222/.7181        &.2040/.2065\\
            \bottomrule
            \end{tabular}
        \end{table}

\section{Conclusion and Future Research} \label{sect:conclusions}

      This paper investigates the vulnerability of CNN classifiers in EEG-based BCI systems. We generate adversarial examples by injecting a jamming module before a CNN classifier to fool it. Three scenarios -- white-box attack, gray-box attack, and black-box attack -- were considered, and separate attack strategies were proposed for each of them. Experiments on three EEG datasets and three CNN classifiers demonstrated the effectiveness of our proposed strategies, i.e., the vulnerability of CNN classifiers in EEG-based BCIs.

      Our future research will:
      \begin{enumerate}
    \item Study the vulnerability of traditional machine learning approaches in BCIs. As shown in Section~\ref{sect:Discussion}, the adversarial examples generated from CNN models may not transfer well to traditional machine learning models, and hence new attack strategies are needed.

      \item Investigate attack strategies to other components of the BCI machine learning model. Fig.~\ref{fig:pipeline} shows that attacks can target different components of a machine learning model. This paper only considered adversarial examples, targeted at the test input. Attacks to the training data, learned model parameters, and the test output, will also be considered.

        \begin{figure}[htbp]   \centering
        \includegraphics[width=.7\linewidth,clip]{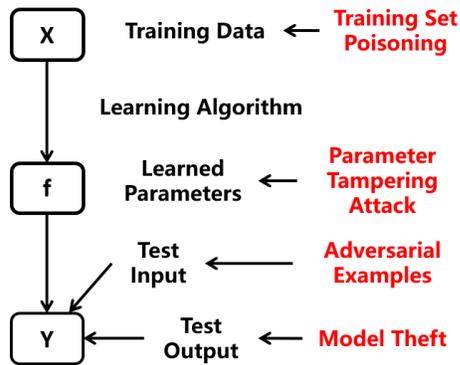}
        \caption{Attack strategies to different components of a machine learning model.} \label{fig:pipeline}
        \end{figure}

      \item Study strategies to defend adversarial attacks on EEG-based BCIs. Multiple defense approaches, e.g., {adversarial training}~\cite{FGSM}, {defensive distillation}~\cite{AdvTransferPapernot2016}, {ensemble adversarial training}~\cite{EnsembleAdvTraining}, and so on~\cite{PGD, Xie, AdvInputTransform}, have been proposed for other application domains. Unfortunately, there has not existed a universal defense approach because it is still unclear theoretically why adversarial examples occur in deep learning. Goodfellow et al. (2014)~\cite{FGSM} believed that adversarial examples exist because of the linearity of deep learning models. Gilmer et al. (2018)~\cite{AdvSphere} argued that adversarial examples occur as a result of the high dimensional geometry of the data manifold. We will investigate the root cause of adversarial examples in EEG classification/regression, and hence develop effective defense strategies for safer BCIs.
      \end{enumerate}

\section*{Acknowledgement}
This research was supported by the National Natural Science Foundation of China Grant 61873321.


\begin{thebibliography}{10}
\providecommand{\url}[1]{#1}
\csname url@samestyle\endcsname
\providecommand{\newblock}{\relax}
\providecommand{\bibinfo}[2]{#2}
\providecommand{\BIBentrySTDinterwordspacing}{\spaceskip=0pt\relax}
\providecommand{\BIBentryALTinterwordstretchfactor}{4}
\providecommand{\BIBentryALTinterwordspacing}{\spaceskip=\fontdimen2\font plus
\BIBentryALTinterwordstretchfactor\fontdimen3\font minus
  \fontdimen4\font\relax}
\providecommand{\BIBforeignlanguage}[2]{{%
\expandafter\ifx\csname l@#1\endcsname\relax
\typeout{** WARNING: IEEEtranS.bst: No hyphenation pattern has been}%
\typeout{** loaded for the language `#1'. Using the pattern for}%
\typeout{** the default language instead.}%
\else
\language=\csname l@#1\endcsname
\fi
#2}}
\providecommand{\BIBdecl}{\relax}
\BIBdecl

\bibitem{FBCSP}
K.~K. Ang, Z.~Y. Chin, H.~Zhang, and C.~Guan, ``Filter bank common spatial
  pattern ({FBCSP}) in brain-computer interface,'' in \emph{Proc. {IEEE} Int'l
  Joint Conf. on Neural Networks}, Hong Kong, China, 2008.

\bibitem{Adv3D}
A.~Athalye, L.~Engstrom, A.~Ilyas, and K.~Kwok, ``Synthesizing robust
  adversarial examples,'' in \emph{Proc. 35th Int'l Conf. on Machine Learning},
  Stockholm, Sweden, Jul. 2018, pp. 284--293.

\bibitem{Barachant2012}
A.~Barachant, S.~Bonnet, M.~Congedo, and C.~Jutten, ``Multiclass
  brain–computer interface classification by riemannian geometry,''
  \emph{IEEE Trans. on Biomedical Engineering}, vol.~59, no.~4, pp. 920--928,
  Apr. 2012.

\bibitem{EEG2Image}
P.~Bashivan, I.~Rish, M.~Yeasin, and N.~Codella, ``Learning representations
  from {EEG} with deep recurrent-convolutional neural networks,'' in
  \emph{Proc. Int'l Conf. on Learning Representations}, San Juan, Puerto Rico,
  May 2016.

\bibitem{DecisionBasedAttack}
W.~Brendel, J.~Rauber, and M.~Bethge, ``Decision-based adversarial attacks:
  Reliable attacks against black-box machine learning models,'' in \emph{Proc.
  Int'l Conf. on Learning Representations}, Vancouver, Canada, May 2018.

\bibitem{AdvPatch}
\BIBentryALTinterwordspacing
T.~B. Brown, D.~Man{\'{e}}, A.~Roy, M.~Abadi, and J.~Gilmer, ``Adversarial
  patch,'' \emph{CoRR}, vol. abs/1712.09665, 2017. [Online]. Available:
  \url{http://arxiv.org/abs/1712.09665}
\BIBentrySTDinterwordspacing

\bibitem{AdvCW}
N.~Carlini and D.~Wagner, ``Towards evaluating the robustness of neural
  networks,'' in \emph{Proc. {IEEE} Symposium on Security and Privacy}.\hskip
  1em plus 0.5em minus 0.4em\relax San Jose, CA: IEEE, May 2017, pp. 39--57.

\bibitem{AdvExamAudio}
\BIBentryALTinterwordspacing
N.~Carlini and D.~A. Wagner, ``Audio adversarial examples: Targeted attacks on
  speech-to-text,'' \emph{CoRR}, vol. abs/1801.01944, 2018. [Online].
  Available: \url{http://arxiv.org/abs/1801.01944}
\BIBentrySTDinterwordspacing

\bibitem{Xception}
F.~Chollet, ``{X}ception: Deep learning with depthwise separable
  convolutions,'' in \emph{Proc. IEEE Conf. on Computer Vision and Pattern
  Recognition}.\hskip 1em plus 0.5em minus 0.4em\relax Honolulu, HI: IEEE, Jul.
  2017, pp. 1800--1807.

\bibitem{P3001988}
L.~Farwell and E.~Donchin, ``Talking off the top of your head: toward a mental
  prosthesis utilizing event-related brain potentials,''
  \emph{Electroencephalography and Clinical Neurophysiology}, vol.~70, no.~6,
  pp. 510--523, 1988.

\bibitem{AdvSphere}
\BIBentryALTinterwordspacing
J.~Gilmer, L.~Metz, F.~Faghri, S.~S. Schoenholz, M.~Raghu, M.~Wattenberg, and
  I.~J. Goodfellow, ``Adversarial spheres,'' \emph{CoRR}, vol. abs/1801.02774,
  2018. [Online]. Available: \url{https://arxiv.org/abs/1801.02774}
\BIBentrySTDinterwordspacing

\bibitem{FGSM}
I.~J. Goodfellow, J.~Shlens, and C.~Szegedy, ``Explaining and harnessing
  adversarial examples,'' in \emph{Proc. Int'l Conf. on Learning
  Representations}, San Diego, CA, May 2015.

\bibitem{BCIIntro}
B.~Graimann, B.~Allison, and G.~Pfurtscheller, \emph{Brain-Computer Interfaces:
  A Gentle Introduction}.\hskip 1em plus 0.5em minus 0.4em\relax Berlin,
  Heidelberg: Springer, 2009, pp. 1--27.

\bibitem{AdvInputTransform}
\BIBentryALTinterwordspacing
C.~Guo, M.~Rana, M.~Ciss{\'{e}}, and L.~van~der Maaten, ``Countering
  adversarial images using input transformations,'' \emph{CoRR}, vol.
  abs/1711.00117, 2017. [Online]. Available:
  \url{https://arxiv.org/abs/1711.00117}
\BIBentrySTDinterwordspacing

\bibitem{EPFLP300}
U.~Hoffmann, J.-M. Vesin, T.~Ebrahimi, and K.~Diserens, ``An efficient
  {P}300-based brain-computer interface for disabled subjects,'' \emph{Journal
  of Neuroscience Methods}, vol. 167, no.~1, pp. 115--125, 2008.

\bibitem{Adam}
D.~P. Kingma and J.~Ba, ``Adam: {A} method for stochastic optimization,'' in
  \emph{Proc. Int'l Conf. on Learning Representations}, Banff, Canada, Apr.
  2014.

\bibitem{Koles1990}
Z.~J. Koles, M.~S. Lazar, and S.~Z. Zhou, ``Spatial patterns underlying
  population differences in the background {EEG},'' \emph{Brain topography},
  vol.~2, no.~4, pp. 275--284, Jun. 1990.

\bibitem{BIM}
A.~Kurakin, I.~J. Goodfellow, and S.~Bengio, ``Adversarial examples in the
  physical world,'' in \emph{Proc. Int'l Conf. on Learning Representations},
  Toulon, France, Apr. 2017.

\bibitem{EEGNet}
V.~J. Lawhern, A.~J. Solon, N.~R. Waytowich, S.~M. Gordon, C.~P. Hung, and
  B.~J. Lance, ``{EEGNet}: a compact convolutional neural network for
  {EEG}-based brain-computer interfaces,'' \emph{Journal of Neural
  Engineering}, vol.~15, no.~5, p. 056013, 2018.

\bibitem{Li2016}
Y.~Li, J.~Pan, J.~Long, T.~Yu, F.~Wang, Z.~Yu, and W.~Wu, ``Multimodal {BCIs}:
  Target detection, multidimensional control, and awareness evaluation in
  patients with disorder of consciousness,'' \emph{Proceedings of the {IEEE}},
  vol. 104, no.~2, pp. 332--352, 2016.

\bibitem{AdvTransferLiu2016}
\BIBentryALTinterwordspacing
Y.~Liu, X.~Chen, C.~Liu, and D.~Song, ``Delving into transferable adversarial
  examples and black-box attacks,'' \emph{CoRR}, vol. abs/1611.02770, 2016.
  [Online]. Available: \url{http://arxiv.org/abs/1611.02770}
\BIBentrySTDinterwordspacing

\bibitem{PGD}
\BIBentryALTinterwordspacing
A.~Madry, A.~Makelov, L.~Schmidt, D.~Tsipras, and A.~Vladu, ``Towards deep
  learning models resistant to adversarial attacks,'' \emph{CoRR}, vol.
  abs/1706.06083, 2017. [Online]. Available:
  \url{http://arXiv.org/abs/1706.06083}
\BIBentrySTDinterwordspacing

\bibitem{ERN}
P.~Margaux, M.~Emmanuel, D.~Sébastien, B.~Olivier, and M.~Jérémie,
  ``Objective and subjective evaluation of online error correction during
  {P}300-based spelling,'' \emph{Advances in Human-Computer Interaction}, vol.
  2012, no. 578295, p.~13, 2012.

\bibitem{Moosavi-Dezfooli2016}
S.-M. Moosavi-Dezfooli, A.~Fawzi, and P.~Frossard, ``Deepfool: a simple and
  accurate method to fool deep neural networks.'' in \emph{Proc. of the {IEEE}
  Conference on Computer Vision and Pattern Recognition}.\hskip 1em plus 0.5em
  minus 0.4em\relax Las Vegas, NV: IEEE, Jun. 2016, pp. 2574--2582.

\bibitem{BCIReview}
L.~F. Nicolas-Alonso and J.~Gomez-Gil, ``Brain computer interfaces, a review,''
  \emph{Sensors}, vol.~12, no.~2, pp. 1211--1279, 2012.

\bibitem{papernot2016transferability}
\BIBentryALTinterwordspacing
N.~Papernot, P.~McDaniel, and I.~Goodfellow, ``Transferability in machine
  learning: from phenomena to black-box attacks using adversarial samples,''
  \emph{CoRR}, vol. abs/1605.07277, 2016. [Online]. Available:
  \url{https://arxiv.org/abs/1605.07277}
\BIBentrySTDinterwordspacing

\bibitem{BlackBoxAttack}
N.~Papernot, P.~McDaniel, I.~Goodfellow, S.~Jha, Z.~B. Celik, and A.~Swami,
  ``Practical black-box attacks against machine learning,'' in \emph{Proc.
  {ACM} Asia Conf. on Computer and Communications Security}.\hskip 1em plus
  0.5em minus 0.4em\relax Abu Dhabi, UAE: ACM, Apr. 2017, pp. 506--519.

\bibitem{AdvTransferPapernot2016}
N.~Papernot, P.~McDaniel, X.~Wu, S.~Jha, and A.~Swami, ``Distillation as a
  defense to adversarial perturbations against deep neural networks,'' in
  \emph{Proc. {IEEE} Symposium on Security and Privacy}.\hskip 1em plus 0.5em
  minus 0.4em\relax San Jose, CA: IEEE, May 2016, pp. 582--597.

\bibitem{MI2001}
G.~Pfurtscheller and C.~Neuper, ``Motor imagery and direct brain-computer
  communication,'' \emph{Proceedings of the IEEE}, vol.~89, no.~7, pp.
  1123--1134, Jul 2001.

\bibitem{Rivet2009}
B.~Rivet, A.~Souloumiac, V.~Attina, and G.~Gibert, ``x{DAWN} algorithm to
  enhance evoked potentials: application to brain–computer interface,''
  \emph{IEEE Trans. on Biomedical Engineering}, vol.~56, no.~8, pp. 2035--2043,
  Aug. 2009.

\bibitem{MNE}
R.~T. Schirrmeister, J.~T. Springenberg, L.~D.~J. Fiederer, M.~Glasstetter,
  K.~Eggensperger, M.~Tangermann, F.~Hutter, W.~Burgard, and T.~Ball, ``Deep
  learning with convolutional neural networks for {EEG} decoding and
  visualization,'' \emph{Human Brain Mapping}, vol.~38, no.~11, pp. 5391--5420,
  2017.

\bibitem{OnePixelAttack}
\BIBentryALTinterwordspacing
J.~Su, D.~V. Vargas, and K.~Sakurai, ``One pixel attack for fooling deep neural
  networks,'' \emph{CoRR}, vol. abs/1710.08864, 2017. [Online]. Available:
  \url{https://arxiv.org/abs/1710.08864}
\BIBentrySTDinterwordspacing

\bibitem{P300Sutton}
S.~Sutton, M.~Braren, J.~Zubin, and E.~R. John, ``Evoked-potential correlates
  of stimulus uncertainty,'' \emph{Science}, vol. 150, no. 3700, pp.
  1187--1188, 1965.

\bibitem{AdvExamSzegedy}
C.~Szegedy, W.~Zaremba, I.~Sutskever, J.~Bruna, D.~Erhan, I.~J. Goodfellow, and
  R.~Fergus, ``Intriguing properties of neural networks,'' in \emph{Proc. Int'l
  Conf. on Learning Representations}, Banff, Canada, Apr. 2014.

\bibitem{DeepLearningMI}
Y.~R. Tabar and U.~Halici, ``A novel deep learning approach for classification
  of {EEG} motor imagery signals,'' \emph{Journal of Neural Engineering},
  vol.~14, no.~1, p. 016003, 2017.

\bibitem{MI4C}
M.~Tangermann, K.-R. Müller, A.~Aertsen, N.~Birbaumer, C.~Braun, C.~Brunner,
  R.~Leeb, C.~Mehring, K.~Miller, G.~Mueller-Putz, G.~Nolte, G.~Pfurtscheller,
  H.~Preissl, G.~Schalk, A.~Schlögl, C.~Vidaurre, S.~Waldert, and
  B.~Blankertz, ``Review of the {BCI} {C}ompetition {IV},'' \emph{Frontiers in
  Neuroscience}, vol.~6, p.~55, 2012.

\bibitem{Tayeb2019}
Z.~Tayeb, J.~Fedjaev, N.~Ghaboosi, C.~Richter, L.~Everding, X.~Qu, Y.~Wu,
  G.~Cheng, and J.~Conradt, ``Validating deep neural networks for online
  decoding of motor imagery movements from {EEG} signals,'' \emph{Sensors},
  vol.~19, no.~1, p. 210, Jan. 2019.

\bibitem{EnsembleAdvTraining}
F.~Tramèr, A.~Kurakin, N.~Papernot, I.~Goodfellow, D.~Boneh, and P.~McDaniel,
  ``Ensemble adversarial training: Attacks and defenses,'' in \emph{Proc. Int'l
  Conf. on Learning Representations}, Vancouver, Canada, May 2018.

\bibitem{AdvTransferSpace}
\BIBentryALTinterwordspacing
F.~Tramèr, N.~Papernot, I.~Goodfellow, D.~Boneh, and P.~McDaniel, ``The space
  of transferable adversarial examples,'' \emph{CoRR}, vol. abs/1704.03453v2,
  2017. [Online]. Available: \url{http://arXiv.org/abs/1704.03453v2}
\BIBentrySTDinterwordspacing

\bibitem{drwuTHMS2017}
D.~Wu, ``Online and offline domain adaptation for reducing {BCI} calibration
  effort,'' \emph{{IEEE} Trans. on Human-Machine Systems}, vol.~47, no.~4, pp.
  550--563, 2017.

\bibitem{drwuSF2018}
D.~Wu, J.-T. King, C.-H. Chuang, C.-T. Lin, and T.-P. Jung, ``Spatial filtering
  for {EEG}-based regression problems in brain-computer interface ({BCI}),''
  \emph{{IEEE} Trans. on Fuzzy Systems}, vol.~26, no.~2, pp. 771--781, 2018.

\bibitem{drwuTFS2017}
D.~Wu, V.~J. Lawhern, S.~Gordon, B.~J. Lance, and C.-T. Lin, ``Driver
  drowsiness estimation from {EEG} signals using online weighted adaptation
  regularization for regression ({OwARR}),'' \emph{{IEEE} Trans. on Fuzzy
  Systems}, vol.~25, no.~6, pp. 1522--1535, 2017.

\bibitem{drwuTNSRE2016}
D.~Wu, V.~J. Lawhern, W.~D. Hairston, and B.~J. Lance, ``Switching {EEG}
  headsets made easy: {Reducing} offline calibration effort using active
  weighted adaptation regularization,'' \emph{{IEEE} Trans. on Neural Systems
  and Rehabilitation Engineering}, vol.~24, no.~11, pp. 1125--1137, 2016.

\bibitem{drwuRG2017}
D.~Wu, V.~J. Lawhern, B.~J. Lance, S.~Gordon, T.-P. Jung, and C.-T. Lin,
  ``{EEG}-based user reaction time estimation using {R}iemannian geometry
  features,'' \emph{{IEEE} Trans. on Neural Systems and Rehabilitation
  Engineering}, vol.~25, no.~11, pp. 2157--2168, 2017.

\bibitem{PKTransAdv}
\BIBentryALTinterwordspacing
L.~Wu, Z.~Zhu, C.~Tai, and W.~E, ``Understanding and enhancing the
  transferability of adversarial examples,'' \emph{CoRR}, vol. abs/1802.09707,
  2018. [Online]. Available: \url{http://arXiv.org/abs/1802.09707}
\BIBentrySTDinterwordspacing

\bibitem{Xie}
C.~Xie, J.~Wang, Z.~Zhang, Z.~Ren, and A.~Yuille, ``Mitigating adversarial
  effects through randomization,'' in \emph{Proc. Int'l Conf. on Learning
  Representations}, Vancouver, Canada, May 2018.

\bibitem{SSVEPSurvey}
D.~Zhu, J.~Bieger, G.~Garcia~Molina, and R.~M. Aarts, ``A survey of stimulation
  methods used in {SSVEP}-based {BCI}s,'' \emph{Computational Intelligence and
  Neuroscience}, p. 702357, 2010.

\end{thebibliography}


\end{document}